\pdfoutput=1

\documentclass[11pt]{article}

\usepackage[final]{acl}

\usepackage{times}
\usepackage{latexsym}
\usepackage{amssymb}
\usepackage{xcolor}
\usepackage{multirow}

\usepackage[T1]{fontenc}

\usepackage[utf8]{inputenc}

\usepackage{microtype}

\usepackage{inconsolata}

\usepackage{graphicx}

\title{PocketLLM: Enabling On-Device Fine-Tuning for Personalized LLMs}

\author{Dan Peng \\
  OPPO Research Institute \\
   Shenzhen, China \\
  \texttt{lepangdan@outlook.com} \\\And
  Zhihui Fu \\
  OPPO Research Institute \\
  Shenzhen, China \\
  \texttt{hzzhzzf@gmail.com} \\\And
  Jun Wang \\
  OPPO Research Institute \\
  Shenzhen, China \\
  \texttt{junwang.lu@gmail.com} \\
  } 

\begin{document}
\maketitle

\begin{abstract}
Recent advancements in large language models (LLMs) have indeed showcased their impressive capabilities. On mobile devices, the wealth of valuable, non-public data generated daily holds great promise for locally fine-tuning personalized LLMs, while maintaining privacy through on-device processing.
However, the constraints of mobile device resources pose challenges to direct on-device LLM fine-tuning, mainly due to the memory-intensive nature of derivative-based optimization required for saving gradients and optimizer states.
To tackle this, we propose employing derivative-free optimization techniques to enable on-device fine-tuning of LLM, even on memory-limited mobile devices.
Empirical results demonstrate that the RoBERTa-large model and OPT-1.3B can be fine-tuned locally on the OPPO Reno 6 smartphone using around 4GB and 6.5GB of memory respectively, using derivative-free optimization techniques. 
This highlights the feasibility of on-device LLM fine-tuning on mobile devices, paving the way for personalized LLMs on resource-constrained devices while safeguarding data privacy.


\end{abstract}

\section{Introduction}
The rapidly evolving field of Large Language Models (LLMs), exemplified by advanced models such as OpenAI’s ChatGPT, marks a substantial breakthrough in artificial intelligence~\cite{cao2023comprehensive}. The implications and benefits of the advancements of LLMs for mobile devices are profound and pervasive. As reported in~\cite{almeida2021smart}~\cite{xu2019first}, the number of deep models incorporated within individual devices is growing rapidly, making
mobile devices are the primary vehicle for AI. 

The continuous generation of private, inaccessible personal data on mobile devices, often diverging from publicly pre-trained LLM distributions, necessitates on-device post-deployment fine-tuning to develop tailored, personalized models while safeguarding data privacy~\cite{li2024personal}.
On-device fine-tuning of personal data locally is an effective solution for model fine-tuning using personal data while ensuring user data privacy, as all data storage and computation occur exclusively on the device without any data leaving it.

Fine-tuning current LLMs on mobile devices with limited resources is challenging due to LLMs' large size, which demands high computational and memory resources. 
Despite some work claims of achieving on-device fine-tuning using various computation-efficient and memory-saving techniques, these implementations are often demonstrated on edge devices like Raspberry Pi~\cite{zhu2023pockengine} rather than on mobile devices such as smartphones and tablets.
Mobile devices, especially smartphones,  more so than other edge devices, generate a substantial amount of highly private and valuable personal data daily due to their extensive usage, holding great potential for enhancing applications by leveraging this data.
However, to the best of our knowledge, there have been no successful on-device fine-tuning implementations on mobile devices to date.

To bridge this gap, our work aims to enable and optimize the fine-tuning of LLMs on resource-constrained mobile devices, particularly smartphones. 
Memory is crucial for determining the feasibility of fine-tuning LLMs on resource-constrained mobile devices locally, while computational capacity and communication bandwidth primarily impact efficiency, particularly latency. Therefore, in this work, our emphasis lies in reducing the memory footprint to make practical fine-tuning on mobile devices feasible, regardless of efficiency concerns. Future efforts are expected to further enhance efficiency.



The substantial memory overhead of LLM fine-tuning arises from the computational and storage demands associated with gradients and optimization states inherent in traditional derivative-based methods.
To tackle this challenge on mobile devices, we propose leveraging derivative-free fine-tuning optimization. This approach aims to reduce the memory footprint during fine-tuning by circumventing the memory-intensive nature of traditional derivative-based methods.
Our experimental results show that we can fine-tune RoBERTa-large and OPT-1.3B on a current off-the-shelf smartphone, OPPO Reno 6, with a memory consumption of around 4GB and 6.5GB, respectively.

We organize our article with the following structure: first, we present related works in~ Section \ref{sec:related_work}, followed by an introduction to our approach (See Section~\ref{sec:appraoch}) and experimental results (See Section~\ref{sec:experiment}). Finally, we conclude with our findings in Section~\ref{sec:conclusion}. Moreover, limitations are discussed in Section~\ref{sec:limitation}. 

\section{Related Works}~\label{sec:related_work}
Numerous studies focus on resource-efficient fine-tuning, which can benefit on-device fine-tuning, categorized into lightweight foundation model design, fine-tuning process optimization, and external resource utilization. Moreover, \cite{wang2024iot} provides a comprehensive survey on integrating LLMs with IoT devices.

\subsection{Design lightweight foundation models}

Employing lightweight foundation models for fine-tuning can reduce computational and memory demands. Techniques such as model pruning \cite{ma2023llm} and quantization \cite{dettmers2022gpt3} are often used to lighten foundation models. However, these compression techniques often degrade the performance of the foundation model, which can further compromise the effectiveness of fine-tuning.
 

\subsection{Optimize fine-tuning processes}

A strand of research is dedicated to optimizing the fine-tuning process to enhance its efficiency in resource consumption. 
\cite{ding2023parameter} minimizes the computational cost of fine-tuning by selectively adjusting a small subset of key model parameters, while~\cite{hu2021lora} achieves this by reformulating updated matrices as products of low-rank ones. 
Despite these approaches reducing computational demands, these approaches still impose a considerable runtime memory burden, making it impractical for memory-constrained mobile devices~\cite{zhang2023lora}. 
On the other hand, 
many works aim to reduce runtime memory usage during fine-tuning by lowering activation memory~\cite{liao2023make}~\cite{zhang2023lora}, using zeroth-order gradient estimator~\cite{malladi2024fine}, or integrating gradient calculation with parameter updates~\cite{lv2023full}. Although memory-efficient, these approaches often suffer from longer running times and may exhibit reduced performance. Our work aligns closely with this line of research. Notably, none of these methods have been implemented on mobile devices, a gap our research addresses.

\subsection{Leverage external resource support}
Another line of work involves offloading some or all of the model's execution to nearby resource-rich edge devices or the cloud~\cite{zhou2019paving}. These approaches leverage external resources to address limitations in resource-constrained scenarios.
However, offloading often entails substantial communication volume, while mobile devices are constrained by limited bandwidth. Moreover, transferring even intermittent data to external devices not owned by the user may pose privacy risks~\cite{he2020attacking}.


\section{Proposed Approach}~\label{sec:appraoch}

\subsection{On-device fine-tuning to ensure privacy}

In this paper, we employ on-device fine-tuning to enable personalized LLM fine-tuning while safeguarding user data privacy. Traditionally, fine-tuning LLMs involves using public data on powerful GPUs hosted by service providers. However, privacy regulations prohibit transferring user personal data to these service providers' servers for the fine-tuning of personalized LLMs~\cite{voigt2017gdpr}. Even with an Edge-Cloud collaboration paradigm ~\cite{yao2022edge}, processing raw data on the user's device to enhance privacy also carries risks, as intermediate data transferred to untrusted clouds could reveal raw data~\cite{he2020attacking}. Our method provides a privacy-preserving solution through on-device fine-tuning, ensuring all computation and storage for fine-tuning remain strictly on the user's device.

\subsection{Critical resource limitations}

Generally, the key resource constraints for fine-tuning on mobile devices
fall into three categories: computational power, memory capacity, and communication bandwidth. 
The computational power affects processing efficiency, with weaker computational power extending fine-tuning time but not necessarily hindering feasibility on mobile devices.
The communication bandwidth does not present a resource constraint in our on-device LLM fine-tuning, without the need for communication, despite serving as a critical bottleneck in offloading settings.
However, memory capacity is critical for the functional feasibility of on-device LLM fine-tuning, as insufficient memory can result in program crashes or out-of-memory errors. Therefore, as an initial step towards on-device LLM fine-tuning, our goal is to minimize the memory footprint to enable LLM fine-tuning on mobile devices.

\subsection{Derivative-free fine-tuning}


In this paper, we propose using derivative-free optimization to locally fine-tune LLMs on mobile devices, mitigating the memory-intensive nature of traditional derivative-based optimization. In derivative-based LLM fine-tuning, such as with SGD and Adam~\cite{kingma2014adam}, the model's states—including parameters, gradients, and optimizer states—constitute the primary part of memory consumption~\cite{ren2021zero}. However, computing gradients and optimizer states is not essential for fine-tuning. The primary objective is to minimize the loss function by identifying optimal parameters. 
In derivative-free techniques, such as evolutionary algorithms and zeroth-order gradient estimators~\cite{spall1992multivariate}, the parameter space is explored by iteratively evaluating the objective function at different points. This approach bypasses the need to compute and store gradients and optimizer states, as required in derivative-based methods, thereby reducing memory usage.

To achieve this, we employ memory-efficient zeroth-order optimization, known as MeZo~\cite{malladi2024fine}, as our chosen method for derivative-free optimization in our work. While MeZo's efficiency is evident on NVIDIA GPUs, its performance on mobile devices remains unexplored, despite its memory-efficient nature. Furthermore, while we utilize MeZo as our implementation, other derivative-free optimization methods are also aligned with our approach.


\section{Experiments}~\label{sec:experiment}
We conducted experiments using MeZo on the OPPO Reno6 smartphone, which has 12GB of memory. Results show MeZo can fine-tune RoBERTa-large and OPT-1.3B using approximately 4GB and 6.5GB of memory, respectively. In contrast, attempting fine-tuning with Adam resulted in an out-of-memory crash. This highlights the memory efficiency of the derivative-free approach, making it viable for fine-tuning LLMs on resource-constrained devices like smartphones.

\subsection{Experimental setting}
We fine-tuned RoBERTa-large on the SST-2 dataset and OPT-1.3B on SuperGLUE tasks, following the MeZo repository~\footnotemark[1]. We conducted all experiments using a commercial off-the-shelf OPPO Reno6 smartphone, employing both the MeZo and Adam fine-tuning methods. Each method runs for 10 steps, ensuring a fair comparison. 

To run MeZo and Adam fine-tuning on Android-based smartphones, we used Termux~\footnotemark[2]~\footnotetext[2]{https://github.com/termux}, a Linux simulation environment for Android. This made it feasible to implement these fine-tuning methods on smartphones, which typically operate on Linux systems with GPUs. 


\subsection{Performance analysis}

\begin{figure}[htbp] 
\centering 
\includegraphics[width=0.5\textwidth]{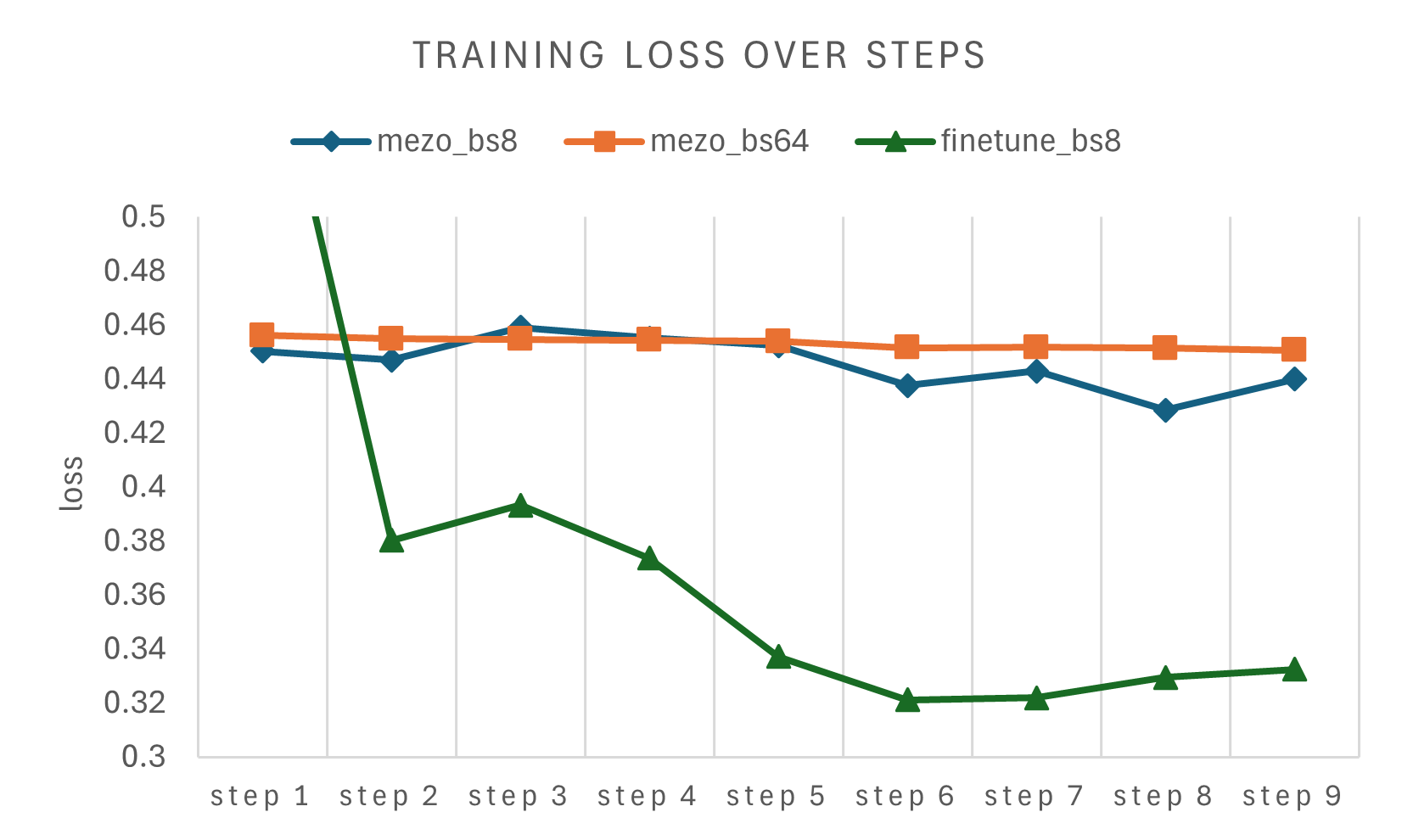} 
\caption{Training loss for  fine-tuning RoBERTa-large using MeZo and Adam fine-tuning.} 
\label{fig:training_loss} 
\end{figure}

We present the training loss during fine-tuning RoBERTa-large using MeZo and Adam fine-tuning on the OPPO Reno 6, as shown in Figure~\ref{fig:training_loss}. We observe that the loss decreases slightly but steadily with MeZo, albeit not as rapidly as with Adam fine-tuning. 
This discrepancy may stem from the estimated gradient's approximation in Mezo, which may not accurately reflect the true gradient and, therefore, the steepest descent direction. This 
demonstrates
the effectiveness of derivative-free fine-tuning, like MeZo, on mobile devices in terms of performance improvement (with decreasing loss), despite its requirement of more steps to converge compared to derivative-based methods.

\subsection{Memory usage analysis}
In Table~\ref{tab:memory_usage }, we compare the memory consumption in fine-tuning RoBERTa-large using MeZo and Adam fine-tuning on the OPPO Reno 6. When using a small batch size of 8, both MeZo and Adam fine-tuning can be conducted on the OPPO Reno 6, with Adam fine-tuning consuming more memory. However, when increasing the batch size to 64, MeZo does not require additional memory, whereas Adam fine-tuning does, exceeding the available memory on the smartphone and resulting in out-of-memory crashes. Further, we fine-tune the larger model OPT-1.3B using MeZo with a memory consumption of about 6.5GB. These all indicate the effectiveness of MeZo for fine-tuning on mobile devices, with regards to memory usage.

We observe that MeZo's memory usage does not significantly increase with batch size, whereas Adam fine-tuning shows a dramatic increase. This is because in derivative-based methods like Adam, activation needs to be saved for gradient computation, and activation linearly increases with batch size. In contrast, derivative-free methods do not require gradient computation or activation saving during optimization, which is an inherent advantage of derivative-free approaches.

\begin{table}[h]
\begin{tabular}{c|l|l}
\hline
\begin{tabular}[c]{@{}c@{}}Memory Usage\\ (GB)\end{tabular} & \multicolumn{1}{c|}{MeZo} & \multicolumn{1}{c}{\begin{tabular}[c]{@{}c@{}}Adam\\ fine-tuning\end{tabular}} \\ 
\hline
\multirow{2}{*}{batch size = 8}& 4.8& 6.5 \\ \cline{2-3} 
 & 4.6 & 6.7\\ \hline
\multirow{2}{*}{batch size = 64}   & 4.0 & OOM  \\ \cline{2-3} 
 & 4.5  & OOM  \\ \hline
\end{tabular}
\caption{Memory usage comparison for fine-tuning RoBERTa-large using MeZo and Adam fine-tuning.}
\label{tab:memory_usage }\end{table}

\subsection{Wall-clock time analysis}
As shown in Table~\ref{tab:wall_clock}, there is no significant difference in per-step training time for RoBERTa-large using MeZo and Adam on the OPPO Reno 6, contradicting the MeZo paper’s claim that MeZo can reduce GPU-hour usage by up to 2× compared to traditional fine-tuning~\cite{malladi2024fine}.
The variance is due to MeZo’s potential to parallelize gradient estimation, unlike backpropagation, which relies on sequential derivative calculations. However, the Reno 6’s limited parallel processing capabilities prevent MeZo from fully utilizing its parallelization potential, resulting in similar per-step training times for both MeZo and Adam, as shown in Table~\ref{tab:wall_clock}.
We also note that parallelization is an inherent vantage of the derivative-free family, extending beyond just MeZo.
 Furthermore, we observe that the per-step training time in MeZo increases with larger batch sizes. This is reasonable because as the batch size increases, the forward pass in MeZo requires more computation.

Moreover, we conduct fine-tuning of the large model OPT-1.3B on the OPPO Reno 6, with a per-step training time of approximately 1800 seconds, which is over 10 times longer than fine-tuning RoBERTa-large. This longer duration is anticipated, given that the parameter size of OPT-1.3B is over 5 times larger than that of RoBERTa-large. Additionally, our experiments show that fine-tuning OPT-1.3B on a single NVIDIA GeForce RTX 3090 GPU takes about 1.99 seconds per step, nearly 1000× faster than on the OPPO Reno 6. This underscores the substantial gap in computational power between mobile devices and GPUs, which are typically used for large model fine-tuning.

\begin{table}[h]
\begin{tabular}{c|l|l}
\hline
\begin{tabular}[c]{@{}c@{}}Training time (s)\\ / per step\end{tabular} & \multicolumn{1}{c|}{MeZo} & \multicolumn{1}{c}{\begin{tabular}[c]{@{}c@{}}Adam\\ fine-tuning\end{tabular}} \\ \hline
\multirow{2}{*}{batch size = 8}  & 97 & 74   \\ \cline{2-3}  & 83   & 85\\ \hline
\multirow{2}{*}{batch size = 64} & 123   & OOM  \\ \cline{2-3}  & 121   & OOM  \\ \hline
\end{tabular}
\caption{Wall-clock time comparison for fine-tuning RoBERTa-large using MeZo and Adam fine-tuning.}
\label{tab:wall_clock}
\end{table}

\section{Conclusions}~\label{sec:conclusion}

We demonstrate that derivative-free optimization allows on-device fine-tuning of LLMs on mobile devices, mitigating the memory constraints of traditional derivative-based methods. Experiments show RoBERTa-large and OPT-1.3B can be fine-tuned on the OPPO Reno 6 using ~4GB and ~6.5GB of memory, respectively. This highlights the advantages of derivative-free optimization for fine-tuning LLMs on resource-constrained mobile devices. 
Further experiments reveal the efficiency gap between smartphones and GPUs, suggesting a need to better utilize hardware capabilities.
Despite these challenges, our successful implementation of fine-tuning LLMs on mobile devices is a significant stride towards personalized models while upholding user data privacy.

\clearpage

\section{Limitations}~\label{sec:limitation}
\subsection{Memory footprint}
While RoBERTa-large and OPT-1.3B have achieved successful fine-tuning with approximately 4GB and 6.5GB of memory respectively, these memory requirements remain too high for typical mobile applications, which often operate within a 1GB memory consumption constraint. It remains crucial to continue minimizing the memory footprint for future implementations.

\subsection{Efficiency of derivative-free family}

Derivative-free optimization methods are often less efficient in determining the optimization direction, which is a strength of derivative-based methods. Therefore, more effective derivative-free methods are needed in future work to reduce the number of steps required for convergence in fine-tuning compared to existing derivative-based methods, thus shortening training times.

\subsection{Adaptation to hardware capabilities}

Despite many flagship mobile devices being equipped with GPUs and even NPUs, which offer powerful computation and parallelization capabilities, the current fine-tuning processes, including our on-device implementation of MeZo, do not fully exploit these hardware capabilities. Derivative-free methods inherently possess parallelization potential, which is currently underutilized. It is crucial to adapt derivative-free methods to fully leverage the powerful computational and parallelization capabilities of current mobile devices.



\subsection{Execution environment}
Our current implementation involves simulating a Linux system using Termux instead of running directly on a mobile device. While beneficial for initial testing, this method serves as a temporary solution and does not accurately reflect performance in a real mobile environment. Specifically, executing programs in Termux may not fully utilize the mobile device's hardware capabilities, potentially leading to suboptimal performance. Additionally, some libraries may be incompatible with Termux, causing issues with the execution of certain algorithms. Moreover, it's important to note that this method does not align with the typical usage scenarios of real users, who interact directly with applications.

A practical approach is to develop native applications that leverage mobile AI frameworks like TensorFlow Lite~\footnotemark[3]~\footnotetext[3]{https://www.tensorflow.org/lite}, empowering developers to integrate LLMs directly into their mobile applications. Future work should strive to deploy on-device fine-tuning algorithms within Android applications. This will facilitate accurate measurement of the algorithm's performance, including efficiency and accuracy, on real-world mobile devices.

\clearpage

\bibliography{acl_latex}

\end{document}